\pdfoutput=1

\documentclass[11pt]{article}
\usepackage[dvipsnames, svgnames, x11names]{xcolor}
\usepackage{emnlp2022}

\usepackage{times}
\usepackage{latexsym}

\usepackage[T1]{fontenc}

\usepackage[utf8]{inputenc}

\usepackage{microtype}

\usepackage{booktabs}
\usepackage{multirow}
\usepackage{amsmath}
\usepackage{amssymb}
\usepackage{xcolor}
\usepackage{pifont}
\usepackage{bbm}

\usepackage{tikz}
\usepackage{subfigure}
\usepackage{pgfplots}
\usepackage{makecell,rotating,diagbox}
\usepackage{verbatim}
\usepackage{colortbl}

\usepackage{algorithm}
\usepackage{algorithmic}
\usepackage{graphicx}
\usepackage{arydshln}
\usepackage{wasysym}

\usetikzlibrary{plotmarks}
\usetikzlibrary{patterns}
\usetikzlibrary{pgfplots.groupplots}

\usetikzlibrary{plotmarks}
\usetikzlibrary{arrows}
\usetikzlibrary{decorations}
\usetikzlibrary{decorations.pathreplacing}
\usetikzlibrary{decorations.markings}
\usetikzlibrary{backgrounds}
\usetikzlibrary{fit}
\usetikzlibrary{positioning}
\usetikzlibrary{calc}
\usetikzlibrary{patterns}
\usetikzlibrary{shadows}
\usetikzlibrary[shapes.multipart]
\usetikzlibrary{pgfplots.groupplots}
\usetikzlibrary{arrows.meta}

\newdimen\legendmargin 
\newdimen\legendwidth 
\newdimen\legendsep 

\definecolor{ugreen}{cmyk}{1,0,1,0.498}
\definecolor{lyyblue}{cmyk}{0.8278,0.3333,0,0.2941}
\definecolor{lyygreen}{cmyk}{0.6813,0,0.725,0.3725}
\definecolor{lyyred}{cmyk}{0,0.8855,0.8767,0.1098}
\definecolor{dblue}{cmyk}{1,0.5487,0,0.5569}
\definecolor{lypurple}{HTML}{e0c2c0}
\definecolor{lygreen}{HTML}{eff67b}
\definecolor{lyblue}{HTML}{d5ddef}
\definecolor{lyblue1}{HTML}{bfe2fe}
\definecolor{lyyellow}{HTML}{fdfab5}
\definecolor{lyyellow1}{HTML}{E1DBE5}
\definecolor{lyypink}{HTML}{ffe0db}
\definecolor{lypink}{HTML}{fecdcb}
\definecolor{lyred}{HTML}{b71a3b}
\definecolor{lygrey}{HTML}{dedfe4}
\definecolor{lyorange}{HTML}{FFEBCD}
\definecolor{zshgreen}{rgb}{0.5607,0.6902,0.5725}
\definecolor{zshorange}{HTML}{fa6d1d}
\definecolor{zshblue}{rgb}{0.6,0.6,1}
\definecolor{zshpink}{rgb}{1,0.6,0.6}

\newcommand{\tab}[1]{Table \ref{#1}}%
\newcommand{\fig}[1]{Fig. \ref{#1}}%
\newcommand{\eqn}[1]{Eq. \ref{#1}}%
\title{Multi-Path Transformer is Better: A Case Study on\\Neural Machine Translation}

\author{}

\author{
  Ye Lin\textsuperscript{1},
  Shuhan Zhou\textsuperscript{1},
  Yanyang Li\textsuperscript{2},
  Anxiang Ma\textsuperscript{1},
  Tong Xiao\textsuperscript{1,3}\thanks{\ \ Corresponding author.},
  Jingbo Zhu\textsuperscript{1,3} \\
  \textsuperscript{1}NLP Lab, School of Computer Science and Engineering, \\
    Northeastern University, Shenyang, China \\
  \textsuperscript{2}The Chinese University of Hong Kong, Hong Kong, China \\
  \textsuperscript{3}NiuTrans Research, Shenyang, China \\
  {\tt \{linye2015,zhoushuhan1997,blamedrlee\}@outlook.com}\\
  {\tt \{maanxiang,xiaotong,zhujingbo\}@mail.neu.edu.cn} \\
}

\begin{document}
\maketitle
\begin{abstract}
    For years the model performance in machine learning obeyed a power-law relationship with the model size.
    For the consideration of parameter efficiency, recent studies focus on increasing model depth rather than width to achieve better performance.
    In this paper, we study how model width affects the Transformer model through a parameter-efficient multi-path structure.
    To better fuse features extracted from different paths, we add three additional operations to each sublayer: a normalization at the end of each path, a cheap operation to produce more features, and a learnable weighted mechanism to fuse all features flexibly.
    Extensive experiments on 12 WMT machine translation tasks show that, with the same number of parameters, the shallower multi-path model can achieve similar or even better performance than the deeper model.
    It reveals that we should pay more attention to the multi-path structure, and there should be a balance between the model depth and width to train a better large-scale Transformer.
\end{abstract}

\section{Introduction}

The large-scale neural network has achieved great success at a wide range of machine learning tasks \cite{DBLP:conf/cvpr/DengDSLL009,DBLP:conf/naacl/DevlinCLT19,DBLP:conf/iclr/ShazeerMMDLHD17}. 
Among them, deep models show great potential to deal with complex problems \cite{DBLP:conf/cvpr/HeZRS16,DBLP:conf/acl/WangLXZLWC19}.
By stacking more layers, deeper models generally perform better than shallower models since they provide more non-linearities to learn more complex transformations \cite{DBLP:journals/corr/Telgarsky15,DBLP:conf/emnlp/LiuLGCH20,DBLP:journals/corr/abs-2008-07772}.
Compared to increasing the model depth, broadening the model width can also benefit the model by providing richer features in a single layer.

There are two ways to broaden the model width: 
1) Scaling the matrix dimensions, such as turning the model configuration from Transformer-base to Transformer-big \cite{DBLP:conf/nips/VaswaniSPUJGKP17}. 
However, both the number of parameters and computational costs will increase quadratically, making such model training and deployment difficult;
2) Adopting the multi-path structure \cite{DBLP:journals/corr/abs-1711-02132}. The expressive power of such models can be improved by fusing abundant features obtained from different paths, and the parameters and computations will only increase linearly with the number of paths.

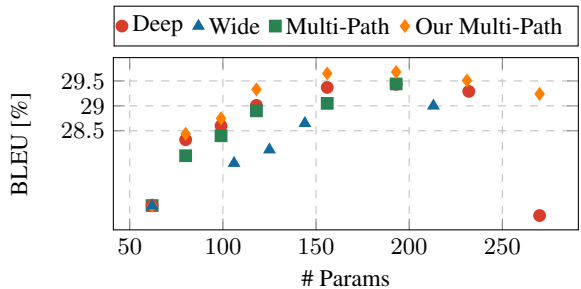
\begin{figure}[t!]
  \hspace{1.3cm}
  \tikz {
      \small
      \legendmargin=0.01\linewidth
      \legendwidth=0.005\linewidth,
      \legendsep=0.008\linewidth
      \coordinate (start) at (0,0);
      \draw[lyyred,thick,postaction={decorate},decoration={markings,mark=at position 0.5 with {\pgfuseplotmark{*}}}] ([xshift=\legendmargin]start.east) -- +(\legendwidth,0) node[black,right] (l1) {Deep};
      \draw[lyyblue,thick,postaction={decorate},decoration={markings,mark=at position 0.5 with {\pgfuseplotmark{triangle*}}}] ([xshift=\legendsep]l1.east) -- +(\legendwidth,0) node[black,right] (l2) {Wide};
      \draw[lyygreen,thick,postaction={decorate},decoration={markings,mark=at position 0.5 with {\pgfuseplotmark{square*}}}] ([xshift=\legendsep]l2.east) -- +(\legendwidth,0) node[black,right] (l3) {Multi-Path};
      \draw[orange,thick,postaction={decorate},decoration={markings,mark=at position 0.5 with {\pgfuseplotmark{diamond*}}}] ([xshift=\legendsep]l3.east) -- +(\legendwidth,0) node[black,right] (l4) {Our Multi-Path};
      \coordinate (end) at ([xshift=\legendmargin+0pt]l4.east);
      \begin{pgfonlayer}{background}
      \node[rectangle,draw,inner sep=0.2pt] [fit = (start) (l1) (l2) (l3) (l4) (end)] {};
      \end{pgfonlayer}
  }
  \\[3pt]
  \centering
  \small
  \begin{tikzpicture}
      \begin{axis}[
          width=1.0\linewidth,
          height=0.502\linewidth,
          yticklabel style={/pgf/number format/fixed,/pgf/number format/precision=1},
          ylabel={BLEU [\%]},
          ylabel near ticks,
          xlabel={\# Params},
          xlabel near ticks,
          enlargelimits=0.1,
          xmajorgrids=true,
          ymajorgrids=true,
          grid style=dashed,
          ytick={28.5,29,29.5},
          every tick label/.append style={font=\small},
          label style={font=\small},
          ylabel style={yshift=5pt},
        ]
          \addplot [
            scatter,
            only marks,
            point meta=explicit symbolic,
            scatter/classes={
              a={mark=*,thick,lyyred},
              b={mark=triangle*,thick,lyyblue},
              c={mark=square*,thick,lyygreen},
              d={mark=diamond*,thick,orange}
            }
          ]table [meta=label] {
            x y label  
            80 28.32 a
            99 28.60 a
            118 29.01 a
            156 29.37 a
            193 29.43 a
            232 29.29 a
            270 26.80 a
            213 29.00 b
            125 28.12 b
            144 28.65 b
            106 27.85 b
            80 28.00 c
            99 28.40 c
            118 28.90 c
            156 29.05 c
            193 29.44 c
            80 28.44 d
            99 28.75 d
            118 29.33 d
            156 29.65 d
            193 29.68 d
            270 29.24 d
            231 29.51 d
            62 27.00 c
            62 27.00 a
            62 27.00 d
            62 27.00 b
          };
        \end{axis}
  \end{tikzpicture}
  \hspace{\fill}
  \caption{Performance (BLEU) vs. the number of parameters (M) on WMT14 En$\rightarrow$De.}
  \label{fig:efficiency}
\end{figure}

The multi-path structure has proven to be quite important in convolutional networks for computer vision tasks \cite{DBLP:journals/corr/abs-2004-08955,DBLP:conf/bmvc/ZagoruykoLLPGCD16}.
But in Transformer, this type of structure has not been widely discussed and applied \cite{DBLP:journals/corr/abs-1711-02132,DBLP:journals/corr/abs-2006-10270}.
In this paper, we continue to study the multi-path structure and adopt a sublayer-level multi-path Transformer model. 
As shown in \fig{fig:efficiency}, the original multi-path models {\small\color{lyygreen}{$\blacksquare$}} significantly outperform wide models {\small\color{lyyblue}{$\blacktriangle$}} that scale the matrix dimensions.
To make better use of the features extracted from different paths, we redesign the multi-path model by introducing three additional operations in each layer: 
1) A normalization at the end of each path for regularization and ease of training; 
2) A cheap operation to produce more features;
3) A learnable weighted mechanism to make the training process more flexible. 

To demonstrate the effectiveness of our method, we test it on the Transformer-base and Transformer-deep configurations.
The experiments are run on 12 WMT machine translation benchmarks, 
including WMT14 English$\leftrightarrow$\{German, French\} and WMT17 English$\leftrightarrow$\{German, Finnish, Latvian, Russian\}.
Experiments on the most widely used English$\rightarrow$German task show that the multi-path Transformer model can achieve 2.65 higher BLEU points with the same depth as the Transformer-base model.
What's more, a shallower multi-path Transformer can achieve 29.68 BLEU points, which is even higher than the 48-layer Transformer-deep model of the same size.
It inspires us that, model width is as important as model depth, instead of indefinitely stacking more layers, we should pay more attention to the multi-path structure.

\section{Background}

\subsection{Transformer}

Transformer is an attention-based encoder-decoder model that has shown promising results in many machine learning tasks \cite{DBLP:conf/nips/VaswaniSPUJGKP17,DBLP:conf/naacl/DevlinCLT19,DBLP:journals/corr/abs-2008-07772}.
It mainly consists of two kinds of structures: the multi-head attention and the feed-forward network.

The multi-head attention computes the attention distribution $A_x$ and then averages the input $X$ by $A_x$. We denote the attention network as $\mathrm{MHA}(\cdot)$:
\begin{align}
    &A_x=\mathrm{SoftMax}(\frac{XW_{q}W_{k}^TX^T}{\sqrt{d}})\label{eqn:self-weight}\\
    &\mathrm{MHA}(X)=A_xXW_{v}\label{eqn:self-sum}
\end{align}
where $X\in\mathbb{R}^{t\times d}$, $t$ is the target sentence length and $d$ is the dimension of the hidden representation, $W_{q},W_{k},W_{v}\in\mathbb{R}^{d\times d}$.

The feed-forward network applies a non-linear transformation to its input $X$. We denote the output as $\mathrm{FFN}(\cdot)$:
\begin{equation}
    \mathrm{FFN}(X)=\mathrm{ReLU}(XW_1+b_1)W_2+b_2\label{eqn:ffn}
\end{equation}
where $W_1\in\mathbb{R}^{d\times 4d}$, $b_1\in\mathbb{R}^{4d}$, $W_2\in\mathbb{R}^{4d\times d}$ and $b_2\in\mathbb{R}^{d}$.

Both the MHA and FFN are coupled with the residual connections \cite{DBLP:conf/cvpr/HeZRS16} and layer normalizations \cite{DBLP:journals/corr/BaKH16}.
For stabilizing deep models training, here we adopt the normalization before layers that has been discussed in \citet{DBLP:conf/acl/WangLXZLWC19}'s work.

\subsection{Multi-Path Transformer}

As a way to enlarge the model capacity, the idea of multi-path networks has been explored widely and has proven to be important in several domains \cite{DBLP:journals/corr/abs-1709-09582,DBLP:conf/aistats/ZhangSS19}.
Among them, 
\citet{DBLP:journals/corr/abs-1711-02132} replace the multi-head attention with multiple self-attentions.
\citet{DBLP:journals/corr/abs-2006-10270} propose MAT, in which the attention layer is the average of multiple independent multi-head attention structures.
The MoE proposes to dynamically choose paths in a very large-scale network \cite{DBLP:conf/iclr/ShazeerMMDLHD17}.

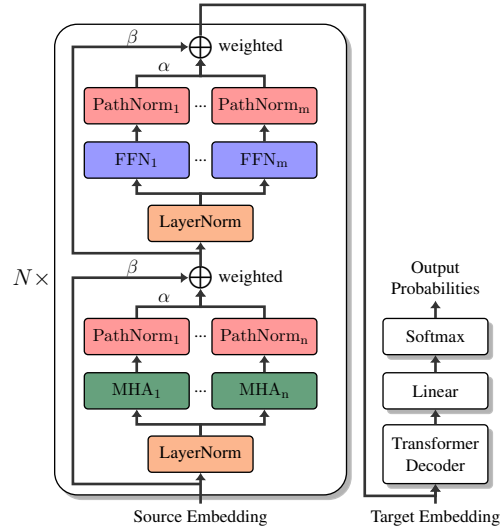
\begin{figure}
    \centering
\tikzstyle{LN} = [rectangle,line width = 0.5pt,rounded corners=0.05cm,minimum width=2.3cm,minimum height=0.8cm,inner sep=0.1cm,draw=black,fill=zshorange!50]
\tikzstyle{MHA} = [rectangle,line width = 0.5pt,rounded corners=0.05cm,minimum width=2.3cm,minimum height=0.8cm,inner sep=0.1cm,draw=black,fill=ugreen!50]
\tikzstyle{PN} = [rectangle,line width = 0.5pt,rounded corners=0.05cm,minimum width=2.3cm,minimum height=0.8cm,inner sep=0.1cm,draw=black,fill=zshpink]
\tikzstyle{FFN} = [rectangle,line width = 0.5pt,rounded corners=0.05cm,minimum width=2.3cm,minimum height=0.8cm,inner sep=0.1cm,draw=black,fill=zshblue]
\tikzstyle{bg} = [line width = 0.5pt,scale=1.3]
\tikzstyle{Decoder} = [rectangle,line width = 0.5pt,rounded corners=0.05cm,minimum width=2.3cm,minimum height=1.3cm,inner sep=0.1cm,draw=black,align=center,fill=white]
\tikzstyle{linear_and_softmax} = [rectangle,line width = 0.5pt,rounded corners=0.05cm,minimum width=2.3cm,minimum height=0.8cm,inner sep=0.1cm,draw=black,fill=white]
\tikzstyle{Background} = [rectangle,line width = 0.5pt,rounded corners=0.3cm,minimum width=6.4cm,minimum height=10.4cm,inner sep=0.1cm,draw=black,fill=white]
\tikzstyle{conect} = [>={LaTeX[width=1.5mm,length=1mm]},line width = 1pt,->,black!80]

\begin{tikzpicture}[node distance = 0,scale = 0.6]
\tikzstyle{every node}=[scale=0.6]
\node(Shadow)[Background,draw=gray!50,fill=gray!60]{};
\node(Background)[Background,above of = Shadow, xshift=-0.1cm,yshift=0.1cm]{};
\node(LN_1)[LN,below of = Background,yshift=-4.3cm]{LayerNorm};
\node(SE)[below of = LN_1, yshift = -1.4cm]{Source Embedding};
\node(MHA_11)[MHA,above of = LN_1,xshift=-1.4cm,yshift=1.4cm,]{$\rm{MHA_1}$};
\node(points1)[above of = LN_1,yshift=1.4cm]{...};
\node(MHA_1n)[MHA,above of = LN_1,xshift=1.4cm,yshift=1.4cm,]{$\rm{MHA_n}$};
\node(PN_11)[PN,above of = MHA_11,yshift=1.2cm,]{$\rm{PathNorm_1}$};
\node(points2)[above of = points1,yshift=1.2cm]{...};
\node(PN_1n)[PN,above of = MHA_1n,yshift=1.2cm,]{$\rm{PathNorm_n}$};
\node(alpha1)[above of = points2,xshift=-0.8cm,yshift=0.85cm]{\large{$\alpha$}};
\node(beta1)[above of = points2,xshift=-1.5cm,yshift=1.5cm]{\large{$\beta$}};
\node(bigoplus1)[above of = points2,yshift=1.3cm,bg]{$\large{\bigoplus}$};
\node(weighted1)[right of = bigoplus1,xshift=1.1cm]{weighted};

\node(N)[left of = weighted1,xshift=-4.8cm]{\Large{$N \times$}};

\node(LN_2)[LN,above of = bigoplus1,yshift=1.2cm]{LayerNorm};
\node(FFN_11)[FFN,above of = LN_2,xshift=-1.4cm,yshift=1.4cm,]{$\rm{FFN_1}$};
\node(points3)[above of = LN_2,yshift=1.4cm]{...};
\node(FFN_1m)[FFN,above of = LN_2,xshift=1.4cm,yshift=1.4cm,]{$\rm{FFN_m}$};
\node(PN_21)[PN,above of = FFN_11,yshift=1.2cm,]{$\rm{PathNorm_1}$};
\node(points4)[above of = points3,yshift=1.2cm]{...};
\node(PN_2m)[PN,above of = FFN_1m,yshift=1.2cm,]{$\rm{PathNorm_m}$};
\node(alpha2)[above of = points4,xshift=-0.8cm,yshift=0.85cm]{\large{$\alpha$}};
\node(beta2)[above of = points4,xshift=-1.5cm,yshift=1.5cm]{\large{$\beta$}};
\node(bigoplus2)[above of = points4,yshift=1.3cm,bg]{$\large{\bigoplus}$};
\node(weighted2)[right of = bigoplus2,xshift=1.1cm]{weighted};

\node(Decoder_shadow)[Decoder,right of = LN_1,xshift=5.25cm,yshift=-0.1cm,draw=gray!50,fill=gray!60]{};
\node(Linear_shadow)[linear_and_softmax,above of = Decoder_shadow,yshift=1.4cm,draw=gray!50,fill=gray!60]{};
\node(Softmax_shadow)[linear_and_softmax,above of = Linear_shadow,yshift=1.2cm,draw=gray!50,fill=gray!60]{};
\node(Decoder)[Decoder,above of = Decoder_shadow,xshift=-0.1cm,yshift=0.1cm]{Transformer\\ Decoder};
\node(TE)[below of = Decoder, yshift = -1.4cm]{Target Embedding};
\node(Linear)[linear_and_softmax,above of = Decoder,yshift=1.4cm]{Linear};
\node(Softmax)[linear_and_softmax,above of = Linear,yshift=1.2cm]{Softmax};
\node(Output)[above of = Softmax,yshift=1.3cm,align=center]{Output\\ Probabilities};

\draw[conect](SE.north)to(LN_1.south);
\draw[conect](LN_1.north)to([yshift=0.25cm]LN_1.north)-|(MHA_11.south);
\draw[conect](LN_1.north)to([yshift=0.25cm]LN_1.north)-|(MHA_1n.south);
\draw[conect](MHA_11.north)to(PN_11.south);
\draw[conect](MHA_1n.north)to(PN_1n.south);
\draw[conect](PN_11.north)to([yshift=0.25cm]PN_11.north)-|([yshift=0.18cm]bigoplus1.south);
\draw[conect](PN_1n.north)to([yshift=0.25cm]PN_1n.north)-|([yshift=0.18cm]bigoplus1.south);
\draw[conect]([yshift=-0.25cm]LN_1.south)to([xshift=-2.8cm,yshift=-0.25cm]LN_1.south)|-([xshift=0.19cm]bigoplus1.west);

\draw[conect]([yshift=-0.18cm]bigoplus1.north)to(LN_2.south);
\draw[conect](LN_2.north)to([yshift=0.25cm]LN_2.north)-|(FFN_11.south);
\draw[conect](LN_2.north)to([yshift=0.25cm]LN_2.north)-|(FFN_1m.south);
\draw[conect](FFN_11.north)to(PN_21.south);
\draw[conect](FFN_1m.north)to(PN_2m.south);
\draw[conect](PN_21.north)to([yshift=0.25cm]PN_21.north)-|([yshift=0.18cm]bigoplus2.south);
\draw[conect](PN_2m.north)to([yshift=0.25cm]PN_2m.north)-|([yshift=0.18cm]bigoplus2.south);
\draw[conect]([yshift=-0.18cm]bigoplus2.north)|-([xshift=3.6cm,yshift=0.5cm]bigoplus2.north)to([xshift=3.6cm,yshift=-10.35cm]bigoplus2.north)-|(Decoder.south);
\draw[conect]([yshift=-0.25cm]LN_2.south)to([xshift=-2.8cm,yshift=-0.25cm]LN_2.south)|-([xshift=0.19cm]bigoplus2.west);

\draw[conect](TE.north)to(Decoder.south);
\draw[conect](Decoder.north)to(Linear.south);
\draw[conect](Linear.north)to(Softmax.south);
\draw[conect](Softmax.north)to(Output.south);
\end{tikzpicture}
    \caption{The architecture of the constructed multi-path Transformer model in this paper.}
    \label{fig:architecture}
\end{figure}

Here, we continue to study the sublayer-level multi-path structure based on the Transformer model.
The multi-path structure is applied both on the multi-head attention and feed-forward network as shown in \fig{fig:architecture},
and the constructed model can be seen as a case of the MoE without dynamic computation.
We discuss that width is an important factor that should not be ignored especially when the model becomes too deep.

\section{Methods}
\label{sec:methods}


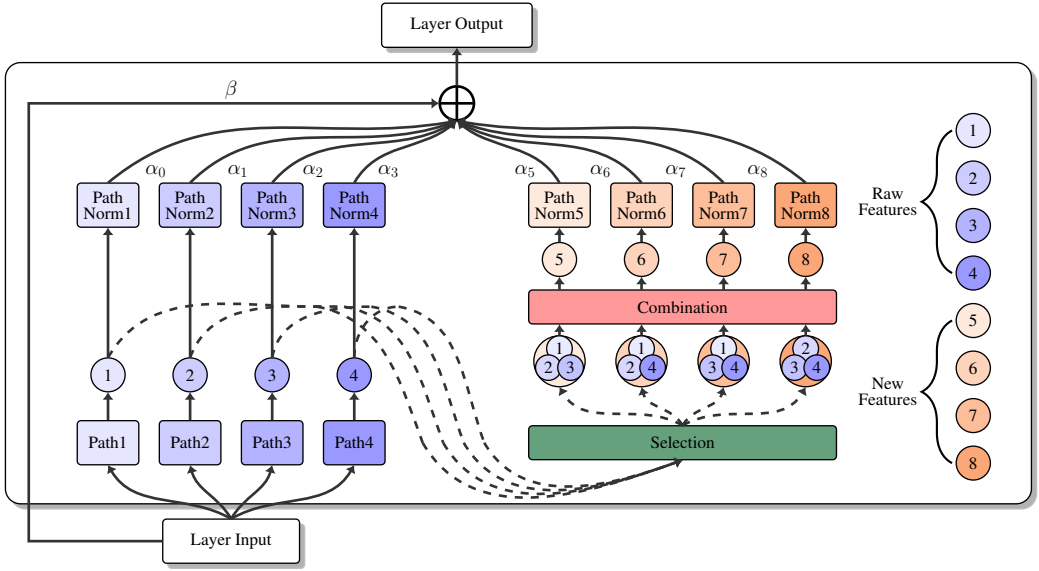
\begin{figure*}
  \centering
\definecolor{zshgreen}{rgb}{0.5607,0.6902,0.5725}
\definecolor{zshorange}{HTML}{fa6d1d}
\definecolor{zshblue}{rgb}{0.6,0.6,1}
\definecolor{zshpink}{rgb}{1,0.6,0.6}
\tikzstyle{Background} = [rectangle,line width = 0.5pt,rounded corners=0.2cm,minimum width=30cm,minimum height=13cm,inner sep=0.1cm,draw=black,fill=white]
\tikzstyle{Path} = [rectangle,line width = 0.5pt,rounded corners=0.05cm,minimum width=1.8cm,minimum height=1.3cm,inner sep=0.1cm,draw=black,fill=zshblue,align=center]
\tikzstyle{input} = [rectangle,line width = 0.5pt,rounded corners=0.05cm,inner sep=0.1cm,minimum width=4cm,minimum height=1.3cm,draw=black,fill=white,align=center]
\tikzstyle{output} = [rectangle,line width = 0.5pt,rounded corners=0.05cm,inner sep=0.1cm,minimum width=4.4cm,minimum height=1.3cm,draw=black,fill=white,align=center]
\tikzstyle{Circle_left} = [circle,line width = 0.5pt,minimum width=1cm,inner sep=0.1cm,draw=black,fill=zshblue]
\tikzstyle{Circle_right} = [circle,line width = 0.5pt,minimum width=1cm,inner sep=0.1cm,draw=black,fill=zshorange!50]
\tikzstyle{Selection} = [rectangle,line width = 0.5pt,rounded corners=0.05cm,minimum width=9cm,minimum height=1cm,inner sep=0.1cm,draw=black,fill=ugreen!50,align=center]
\tikzstyle{Big_circle} = [circle,line width = 0.5pt,minimum width=1.5cm,inner sep=0.1cm,draw=black,fill=zshorange!50]
\tikzstyle{Little_circle1} = [circle,line width = 0.5pt,minimum width=0.75cm,inner sep=0.1cm,draw=black,fill=zshblue!20]
\tikzstyle{Little_circle2} = [circle,line width = 0.5pt,minimum width=0.75cm,inner sep=0.1cm,draw=black,fill=zshblue!40]
\tikzstyle{Little_circle3} = [circle,line width = 0.5pt,minimum width=0.75cm,inner sep=0.1cm,draw=black,fill=zshblue!60]
\tikzstyle{Little_circle4} = [circle,line width = 0.5pt,minimum width=0.75cm,inner sep=0.1cm,draw=black,fill=zshblue]
\tikzstyle{Combination} = [rectangle,line width = 0.5pt,rounded corners=0.05cm,minimum width=9cm,minimum height=1cm,inner sep=0.1cm,draw=black,fill=zshpink,align=center]
\tikzstyle{conect} = [>={LaTeX[width=1.5mm,length=1mm]},line width = 1pt,->,black!80]

\begin{tikzpicture}[node distance = 0,scale = 0.45]
\tikzstyle{every node}=[scale=0.45]
\node(shadow)[Background,draw=gray!50,fill=gray!60]{};
\node(bg)[Background,above of = shadow,xshift=-0.15cm,yshift=0.15cm]{};
\node(Path1)[Path,right of = bg,xshift=-12cm, yshift=-4.7cm,fill=zshblue!25]{\Large{Path1}};
\node(Path2)[Path,right of = Path1,xshift=2.4cm,fill=zshblue!50]{\Large{Path2}};
\node(Path3)[Path,right of = Path2,xshift=2.4cm,fill=zshblue!75]{\Large{Path3}};
\node(Path4)[Path,right of = Path3,xshift=2.4cm,fill=zshblue]{\Large{Path4}};
\node(input_shadow)[input,right of = Path2,xshift=1.35cm,yshift=-3.05cm,align=center,draw=gray!50,fill=gray!60]{};
\node(input)[input,right of = Path2,xshift=1.2cm,yshift=-2.9cm,align=center]{\Large{Layer Input}};
\node(Circle1)[Circle_left,above of = Path1,yshift=2cm,fill=zshblue!25]{\Large{1}};
\node(Circle2)[Circle_left,above of = Path2,yshift=2cm,fill=zshblue!50]{\Large{2}};
\node(Circle3)[Circle_left,above of = Path3,yshift=2cm,fill=zshblue!75]{\Large{3}};
\node(Circle4)[Circle_left,above of = Path4,yshift=2cm,fill=zshblue]{\Large{4}};
\node(PathNorm1)[Path,above of = Circle1,yshift=5cm,fill=zshblue!25]{\Large{Path}\\ \Large{Norm1}};
\node(PathNorm2)[Path,above of = Circle2,yshift=5cm,fill=zshblue!50]{\Large{Path}\\ \Large{Norm2}};
\node(PathNorm3)[Path,above of = Circle3,yshift=5cm,fill=zshblue!75]{\Large{Path}\\ \Large{Norm3}};
\node(PathNorm4)[Path,above of = Circle4,yshift=5cm,fill=zshblue]{\Large{Path}\\ \Large{Norm4}};

\node(beta)[above of = PathNorm2,xshift=1.2cm,yshift=3.4cm]{\LARGE{$\beta$}};
\node(alpha_0)[above of = PathNorm1,xshift=1.4cm,yshift=1cm]{\LARGE{$\alpha_0$}};
\node(alpha_1)[above of = PathNorm2,xshift=1.4cm,yshift=1cm]{\LARGE{$\alpha_1$}};
\node(alpha_2)[above of = PathNorm3,xshift=1.2cm,yshift=1cm]{\LARGE{$\alpha_2$}};
\node(alpha_3)[above of = PathNorm4,xshift=1cm,yshift=1cm]{\LARGE{$\alpha_3$}};

\node(bigoplus)[circle,above of = PathNorm4,xshift=3cm,yshift=3cm,minimum width=1cm,line width=1pt,draw=black]{};
\draw[-,line width=1pt,](bigoplus.west)to(bigoplus.east);
\draw[-,line width=1pt,](bigoplus.north)to(bigoplus.south);
\node(output_shadow)[output,above of = bigoplus,xshift=0.15cm,yshift=2.15cm,align=center,draw=gray!50,fill=gray!60]{};
\node(output)[output,above of = bigoplus,yshift=2.3cm,align=center]{\Large{Layer Output}};

\node(Selection)[Selection,right of = Path4,xshift=9.6cm]{\Large{Selection}};
\node(Big_circle1)[Big_circle,above of = Selection,xshift=-3.6cm,yshift=2.4cm,fill=zshorange!15]{};
\node(litte_circle11)[Little_circle1,above of = Big_circle1,xshift=0cm,yshift=0.375cm]{\Large{1}};
\node(litte_circle12)[Little_circle2,above of = Big_circle1,xshift=-0.32476cm,yshift=-0.1875cm]{\Large{2}};
\node(litte_circle13)[Little_circle3,above of = Big_circle1,xshift=0.32476cm,yshift=-0.1875cm]{\Large{3}};
\node(Big_circle2)[Big_circle,above of = Selection,xshift=-1.2cm,yshift=2.4cm,fill=zshorange!30]{};
\node(litte_circle21)[Little_circle1,above of = Big_circle2,xshift=0cm,yshift=0.375cm]{\Large{1}};
\node(litte_circle22)[Little_circle2,above of = Big_circle2,xshift=-0.32476cm,yshift=-0.1875cm]{\Large{2}};
\node(litte_circle23)[Little_circle4,above of = Big_circle2,xshift=0.32476cm,yshift=-0.1875cm]{\Large{4}};
\node(Big_circle3)[Big_circle,above of = Selection,xshift=1.2cm,yshift=2.4cm,fill=zshorange!45]{};
\node(litte_circle31)[Little_circle1,above of = Big_circle3,xshift=0cm,yshift=0.375cm]{\Large{1}};
\node(litte_circle32)[Little_circle3,above of = Big_circle3,xshift=-0.32476cm,yshift=-0.1875cm]{\Large{3}};
\node(litte_circle33)[Little_circle4,above of = Big_circle3,xshift=0.32476cm,yshift=-0.1875cm]{\Large{4}};
\node(Big_circle4)[Big_circle,above of = Selection,xshift=3.6cm,yshift=2.4cm,fill=zshorange!60]{};
\node(Combination)[Combination,above of = Selection,yshift=4cm]{\Large{Combination}};
\node(litte_circle41)[Little_circle2,above of = Big_circle4,xshift=0cm,yshift=0.375cm]{\Large{2}};
\node(litte_circle42)[Little_circle3,above of = Big_circle4,xshift=-0.32476cm,yshift=-0.1875cm]{\Large{3}};
\node(litte_circle43)[Little_circle4,above of = Big_circle4,xshift=0.32476cm,yshift=-0.1875cm]{\Large{4}};
\node(Circle5)[Circle_right,above of = Big_circle1,yshift=3cm,fill=zshorange!15]{\Large{5}};
\node(Circle6)[Circle_right,above of = Big_circle2,yshift=3cm,fill=zshorange!30]{\Large{6}};
\node(Circle7)[Circle_right,above of = Big_circle3,yshift=3cm,fill=zshorange!45]{\Large{7}};
\node(Circle8)[Circle_right,above of = Big_circle4,yshift=3cm,fill=zshorange!60]{\Large{8}};
\node(PathNorm5)[Path,above of = Circle5,yshift=1.6cm,fill=zshorange!15]{\Large{Path}\\ \Large{Norm5}};
\node(PathNorm6)[Path,above of = Circle6,yshift=1.6cm,fill=zshorange!30]{\Large{Path}\\ \Large{Norm6}};
\node(PathNorm7)[Path,above of = Circle7,yshift=1.6cm,fill=zshorange!45]{\Large{Path}\\ \Large{Norm7}};
\node(PathNorm8)[Path,above of = Circle8,yshift=1.6cm,fill=zshorange!60]{\Large{Path}\\ \Large{Norm8}};

\node(alpha_7)[above of = PathNorm8,xshift=-1.4cm,yshift=1cm]{\LARGE{$\alpha_8$}};
\node(alpha_6)[above of = PathNorm7,xshift=-1.4cm,yshift=1cm]{\LARGE{$\alpha_7$}};
\node(alpha_5)[above of = PathNorm6,xshift=-1.2cm,yshift=1cm]{\LARGE{$\alpha_6$}};
\node(alpha_4)[above of = PathNorm5,xshift=-1cm,yshift=1cm]{\LARGE{$\alpha_5$}};

\node(Circle_8)[Circle_right,right of = Selection,xshift=8.5cm,yshift=-0.6cm,fill=zshorange!60]{\Large{8}};
\node(Circle_7)[Circle_right,above of = Circle_8,yshift=1.4cm,fill=zshorange!45]{\Large{7}};
\node(new)[above of = Circle_7,xshift=-2.5cm,yshift=0.7cm,align=center]{\Large{New}\\ \Large{Features}};
\node(Circle_6)[Circle_right,above of = Circle_7,yshift=1.4cm,fill=zshorange!30]{\Large{6}};
\node(Circle_5)[Circle_right,above of = Circle_6,yshift=1.4cm,fill=zshorange!15]{\Large{5}};
\draw[decorate,decoration={brace,raise=0.05cm,amplitude=0.4cm},thick] (Circle_8.west) -- (Circle_5.west);
\node(Circle_4)[Circle_left,above of = Circle_5,yshift=1.4cm,fill=zshblue]{\Large{4}};
\node(Circle_3)[Circle_left,above of = Circle_4,yshift=1.4cm,fill=zshblue!75]{\Large{3}};
\node(raw)[above of = Circle_3,xshift=-2.5cm,yshift=0.7cm,align=center]{\Large{Raw}\\ \Large{Features}};
\node(Circle_2)[Circle_left,above of = Circle_3,yshift=1.4cm,fill=zshblue!50]{\Large{2}};
\node(Circle_1)[Circle_left,above of = Circle_2,yshift=1.4cm,fill=zshblue!25]{\Large{1}};
\draw[decorate,decoration={brace,raise=0.05cm,amplitude=0.4cm},thick] (Circle_4.west) -- (Circle_1.west);

\draw[conect,in=-60,out=150] (input.north) to (Path1.south);
\draw[conect,in=-75,out=120] (input.north) to (Path2.south);
\draw[conect,in=-105,out=60] (input.north) to (Path3.south);
\draw[conect,in=-120,out=30] (input.north) to (Path4.south);

\draw[conect] (Path1.north) to (Circle1.south);
\draw[conect] (Path2.north) to (Circle2.south);
\draw[conect] (Path3.north) to (Circle3.south);
\draw[conect] (Path4.north) to (Circle4.south);

\draw[conect] (Circle1.north) to (PathNorm1.south);
\draw[conect] (Circle2.north) to (PathNorm2.south);
\draw[conect] (Circle3.north) to (PathNorm3.south);
\draw[conect] (Circle4.north) to (PathNorm4.south);

\draw[conect,dashed,in=-60,out=150] (Selection.north) to (Big_circle1.south);
\draw[conect,dashed,in=-75,out=120] (Selection.north) to (Big_circle2.south);
\draw[conect,dashed,in=-105,out=60] (Selection.north) to (Big_circle3.south);
\draw[conect,dashed,in=-120,out=30] (Selection.north) to (Big_circle4.south);

\draw[conect,-,dashed,in=180,out=60] (Circle1.north) to ([xshift=6.1cm,yshift=1.5cm]Circle1.north);
\draw[conect,-,dashed,in=180,out=70] (Circle2.north) to ([xshift=6.9cm,yshift=1.5cm]Circle1.north);
\draw[conect,-,dashed,in=180,out=80] (Circle3.north) to ([xshift=7.7cm,yshift=1.5cm]Circle1.north);
\draw[conect,-,dashed,in=180,out=90] (Circle4.north) to ([xshift=8.5cm,yshift=1.5cm]Circle1.north);

\draw[conect,-,dashed,in=110,out=-10] ([xshift=6.1cm,yshift=1.5cm]Circle1.north) to ([xshift=-7.6cm]Selection);
\draw[conect,-,dashed,in=110,out=-10] ([xshift=6.9cm,yshift=1.5cm]Circle1.north) to ([xshift=-7cm]Selection);
\draw[conect,-,dashed,in=110,out=-10] ([xshift=7.7cm,yshift=1.5cm]Circle1.north) to ([xshift=-6.4cm]Selection);
\draw[conect,-,dashed,in=110,out=-10] ([xshift=8.5cm,yshift=1.5cm]Circle1.north) to ([xshift=-5.8cm]Selection);

\draw[conect,dashed,in=-160,out=-60] ([xshift=-7.6cm]Selection.center) to  (Selection.south);
\draw[conect,dashed,in=-160,out=-60] ([xshift=-7cm]Selection.center) to  (Selection.south);
\draw[conect,dashed,in=-160,out=-60] ([xshift=-6.4cm]Selection.center) to  (Selection.south);
\draw[conect,dashed,in=-160,out=-60] ([xshift=-5.8cm]Selection.center) to  (Selection.south);

\draw[conect] (Big_circle1.north) to ([xshift=-3.6cm]Combination.south);
\draw[conect] (Big_circle2.north) to ([xshift=-1.2cm]Combination.south);
\draw[conect] (Big_circle3.north) to ([xshift=1.2cm]Combination.south);
\draw[conect] (Big_circle4.north) to ([xshift=3.6cm]Combination.south);

\draw[conect] ([xshift=-3.6cm]Combination.north) to (Circle5.south);
\draw[conect] ([xshift=-1.2cm]Combination.north) to (Circle6.south);
\draw[conect] ([xshift=1.2cm]Combination.north) to (Circle7.south);
\draw[conect] ([xshift=3.6cm]Combination.north) to (Circle8.south);

\draw[conect] (Circle5.north) to (PathNorm5.south);
\draw[conect] (Circle6.north) to (PathNorm6.south);
\draw[conect] (Circle7.north) to (PathNorm7.south);
\draw[conect] (Circle8.north) to (PathNorm8.south);

\draw[conect,in=-170,out=40] (PathNorm1.north) to (bigoplus.south);
\draw[conect,in=-160,out=45] (PathNorm2.north) to (bigoplus.south);
\draw[conect,in=-150,out=50] (PathNorm3.north) to (bigoplus.south);
\draw[conect,in=-140,out=60] (PathNorm4.north) to (bigoplus.south);
\draw[conect,in=-10,out=140] (PathNorm8.north) to (bigoplus.south);
\draw[conect,in=-20,out=135] (PathNorm7.north) to (bigoplus.south);
\draw[conect,in=-30,out=130] (PathNorm6.north) to (bigoplus.south);
\draw[conect,in=-40,out=120] (PathNorm5.north) to (bigoplus.south);

\draw[conect] (bigoplus.north) to (output.south);
\draw[conect] (input.west) -| ([xshift=-12cm]bigoplus.west)to(bigoplus.west);
\end{tikzpicture}
  \caption{A running example of the process of generating more features from cheap operations.}
  \label{fig:features}
\end{figure*}

\subsection{PathNorm and The Weighted Mechanism}
\label{sec:multi-path}

\fig{fig:features} shows the architecture of the multi-path Transformer model constructed in this paper.
In the implementation, we adopt the normalization before layers because it has proven to be more robust to deep models than the normalization after layers \cite{DBLP:conf/iclr/BaevskiA19,DBLP:conf/icml/XiongYHZZXZLWL20,DBLP:journals/corr/abs-1910-05895}.
In this model, different paths are split after each layer normalization and fused at a sublayer level.
To better fuse features extracted from different paths, three additional operations are proposed in this paper.
In this section, we will introduce two of these three operations, the other one will be introduced in Section \ref{sec:features}.

\textbf{PathNorm.}
As shown in \fig{fig:architecture}, an additional normalization (named $\mathrm{PathNorm}$) is introduced at the end of each multi-head attention (MHA) or feed-forward network (FFN).
Different from \citet{DBLP:journals/corr/abs-2110-09456}'s work,
the proposed $\mathrm{PathNorm}$ aims to bring the magnitudes of output distributions closer, which we think is more conducive to the fusion of different paths.
When the number of paths becomes relatively large, it also plays a role in regularization and ensures the stability of the model training.

\textbf{The Weighted Mechanism.}
To enable the model to learn how to combine paths on its own, a learnable weighted mechanism is introduced.
As shown in \fig{fig:architecture}, learnable weights $\alpha$ are added on all model paths, and the residual connections surrounding layers are also equipped with learnable weights $\beta$.
By adopting this strategy, the model can automatically distinguish which part is more important and the training process will be more flexible.

For this multi-path Transformer model, we can write the output of multi-head attention or feed-forward network as:

\begin{equation}
    Y=\beta X+\sum_{i=1}^n\alpha_i\mathrm{Path}_i(X)
\end{equation}
where $X$ denotes the layer input, $Y$ denotes the layer output, $n$ denotes the total number of paths.
$\alpha$ and $\beta$ are respectively learnable weights added on the model paths and residual connections.
We denote the computation of the multi-head attention in \eqn{eqn:self-sum} as $\mathrm{MHA}(\cdot)$ and the computation of the feed-forward network in \eqn{eqn:ffn} as $\mathrm{FFN}(\cdot)$.
In the multi-head attention or feed-forward network, each path can be computed as:
\begin{align}
    \mathrm{Path}(X)&=\mathrm{PathNorm}(\mathrm{Func}(\mathrm{LN}(X)))\label{eqn:path}\\
    \mathrm{Func}(X)&=\mathrm{MHA}(X) \ or \ \mathrm{FFN}(X)\label{eqn:func}
\end{align} 
where $\mathrm{PathNorm}$ is the normalization added after the computation of each multi-head attention or feed-forward layer.
$\mathrm{LN}$ is the layer normalization.

\subsection{More Features from Cheap Operations}
\label{sec:features}

With the increasing number of paths, the model tends to get better performance, but the number of parameters and computational costs will also increase correspondingly. 
What's worse, one model with too many paths will be hard to train because of much more GPU memory resources consuming.

To solve the above-mentioned problem, here we propose to generate more features from the existing ones through a cheap operation.
This method can help the multi-path model achieve better performance with almost negligible additional computational costs, and it has no effect on the overall parameters.
Specifically, here we adopt a "selection then combination" strategy.
In the example of \fig{fig:features}, Paths $1\sim 4$ denote paths of the current Transformer layer, the features generated from these paths are called "raw features", and the features further generated by "raw features" are called "new features".
This process can be divided into the listed two steps.

\textbf{Selection.}
Firstly, we need to select a few paths for the next combination operation.
Since we want to get different features, the selection must be without repetition.
Since different paths are independent, so the selection order does not matter.
It should be noted that if each path is selected too many times, it will weaken the feature diversities.
While, if each path is chosen only a small number of times, there will be fewer benefits.
Here we select $n-1$ paths from the total $n$ paths once time, until all subsets of paths that meet this condition are selected without repetition.
From $n$ paths, there will be $C_n^{n-1}=n$ subsets of paths, and the corresponding number of newly generated features will be $n$.
Through this selection strategy, there will be a balance between the rawly existing features and the newly generated features.

\textbf{Combination.}
Since we have obtained $n$ subsets of paths from the selection operation, we need to combine paths from the same subset to generate new features.
To compute the combination result, here we adopt a simple average operation.
Specifically, we average the outputs of different paths in each attention or feed-forward network which have been computed in \eqn{eqn:func}.
Suppose the number of paths is set to $n$, then the number of paths in one subset is $k=n-1$, the average operation can be denoted as below:
\begin{equation}
    \mathrm{AVG}(X)=\frac{\sum_{i=1}^{k}{\mathrm{Func}}_i(X)}{k}
\end{equation}

After the combination operation to produce $n$ new features, we add additional normalizations as described in Section \ref{sec:multi-path}.
Different from the previous description, here we add $\mathrm{PathNorm}$ on both the "raw features" and "new features".
Besides, learnable weights $\alpha$ and $\beta$ are also added for weighting these two kinds of features as shown in \fig{fig:features}.

\textbf{Efficiency.}
Considering the parameter efficiency, although we need to introduce additional normalizations with twice the number of model paths, it has little effect on the total number of parameters since the parameters in each normalization are very limited.
As for the computation efficiency, because the average operation is quite lightweight and the dimension of the sublayer output is relatively small, the combination operation will only have a small impact on the overall training efficiency.
Since we only experiment on the Transformer encoder, it has nearly no impact on the inference efficiency.

\subsection{The Initialization of $\alpha$ and $\beta$}
In Section \ref{sec:multi-path}, we have introduced the learnable weights $\alpha$ and $\beta$ for the path outputs and residual connections respectively.
Here we introduce how to initialize $\alpha$ and $\beta$ in this work.
Since the result of one path coupled with $\mathrm{PathNorm}$ 
follows the normal distribution with mean 0 and variance 1, to make the sum of multiple paths approximately equal to the standard normal distribution, here we set $\alpha=\frac{1}{\sqrt{2n}}$, where $n$ is the number of "raw features" in the current layer. 
To balance the residual connections and paths in the initial training stage, we set $\beta=1$ in all sublayers.

\begin{table}[t!]
  \centering
  \small
  \renewcommand\tabcolsep{2.5pt}
  \renewcommand\arraystretch{1.3}
  \begin{tabular}{l|l|r|r|r|r|r|r}
      \hline
      \multicolumn{1}{c|}{\multirow{2}*{Source}} & 
      \multicolumn{1}{c|}{\multirow{2}*{Task}} & 
      \multicolumn{2}{c|}{Train} & 
      \multicolumn{2}{c|}{Valid} & 
      \multicolumn{2}{c}{Test} \\
      \cline{3-8}
      & &
      \multicolumn{1}{c|}{sent.} & 
      \multicolumn{1}{c|}{word} &
      \multicolumn{1}{c|}{sent.} & 
      \multicolumn{1}{c|}{word} &
      \multicolumn{1}{c|}{sent.} &
      \multicolumn{1}{c}{word} \\
      \hline
      \multirow{2}*{WMT14} & En$\leftrightarrow$De & 4.5M & 220M & 3000 & 110K & 3003 & 114K \\
      \cline{2-8}
      & En$\leftrightarrow$Fr & 35M & 2.2B & 26K & 1.7M & 3003 & 155K \\
      \cline{1-8}
      \multirow{5}*{WMT17} & En$\leftrightarrow$De & 5.9M & 276M & 8171 & 356K & 3004 & 128K \\
      \cline{2-8}
      & En$\leftrightarrow$Fi & 2.6M & 108M & 8870 & 330K & 3002 & 110K \\
      \cline{2-8}
      & En$\leftrightarrow$Lv & 4.5M & 115M & 2003 & 90K & 2001 & 88K \\
      \cline{2-8}
      & En$\leftrightarrow$Ru & 25M & 1.2B & 8819 & 391K & 3001 & 132K \\
      \hline
  \end{tabular}
  \caption{Data statistics (\# of sentences and \# of words).}
  \label{tab:data}
\end{table}

\section{Experiments}

\begin{table*}[t!]
  \centering
  \small
  \renewcommand\arraystretch{1.11}
  \setlength{\tabcolsep}{0.33cm}
  \begin{tabular}{lr|r|c|c|c|c|c|c|c}
  \hline
  \multicolumn{1}{l}{System} &
  &
  Params &
  \multicolumn{1}{c|}{Depth} &
  \multicolumn{1}{c|}{Path} &
  \multicolumn{1}{c|}{Test} &
  \multicolumn{1}{c|}{$\mathrm{\Delta}_{\text{BLEU}}$} &
  \multicolumn{1}{c|}{Valid} &
  \multicolumn{1}{c|}{$\mathrm{\Delta}_{\text{BLEU}}$} &
  \multicolumn{1}{c}{$\mathrm{\Delta}_{\text{Average}}$} \\
  \hline
  \multirow{1}{*}{Transformer-base} & & 62M & 6 & 1 & 27.00 & +0.00 & 25.88 & +0.00 & +0.00 \\
  \hline
  \hline
  \multirow{1}{*}{\textbf{Deep12} \ding{51}} &  & \textbf{80M} & \textbf{12} & \textbf{1} & 28.32 & +1.32 & \textbf{26.65} & \textbf{+0.77} & \textbf{+1.05} \\
  \hdashline[5pt/2.5pt]
  \multirow{1}{*}{Multi-Path2} & & 80M & 6 & 2 & 28.00 & +1.00 & 26.22 & +0.34 & +0.67 \\
  \multirow{1}{*}{\quad + Ours} &  & 80M & 6 & 2 & \textbf{28.44} & \textbf{+1.44} & 26.43 & +0.55 & +1.00 \\
  \hline
  \hline
  \multirow{1}{*}{Deep24} &  & 118M & 24 & 1 & 29.01 & +2.01 & 26.94 & +1.06 & +1.54 \\
  \hdashline[5pt/2.5pt]
  \multirow{1}{*}{Multi-Path2} & & 118M & 12 & 2 & 28.90 & +1.90 & 26.67 & +0.79 & +1.35 \\
  \multirow{1}{*}{\quad \textbf{+ Ours} \ding{51}} &  & \textbf{118M} & \textbf{12} & \textbf{2} & \textbf{29.33} & \textbf{+2.33} & \textbf{27.23} & \textbf{+1.35} & \textbf{+1.84} \\
  \hdashline[5pt/2.5pt]
  \multirow{1}{*}{Multi-Path4} & & 118M & 6 & 4 & 28.25 & +1.25 & 26.18 & +0.30 & +0.78 \\
  \multirow{1}{*}{\quad + Ours} &  & 118M & 6 & 4 & 29.04 & +2.04 & 26.80 & +0.92 & +1.48 \\
  \multirow{1}{*}{\quad + More Features} &  & 118M & 6 & 4 & 29.19 & +2.19 & 26.96 & +1.08 & +1.64 \\
  \hline
  \hline
  \multirow{1}{*}{Deep36} &  & 156M & 36 & 1 & 29.37 & +2.37 & \textbf{27.14} & \textbf{+1.26} & +1.82 \\ 
  \hdashline[5pt/2.5pt]
  \multirow{1}{*}{Multi-Path3} & & 156M & 12 & 3 & 29.05 & +2.05 & 26.69 & +0.81 & +1.43 \\
  \multirow{1}{*}{\quad + Ours} &  & 156M & 12 & 3 & 29.08 & +2.08 & 26.93 & +1.05 & +1.57 \\
  \multirow{1}{*}{\quad + More Features} &  & 156M & 12 & 3 & 29.20 & +2.20 & 27.05 & +1.17 & +1.69 \\   
  \hdashline[5pt/2.5pt]
  \multirow{1}{*}{Multi-Path6} & & 156M & 6 & 6 & 28.87 & +1.87 & 26.58 & +0.70 & +1.29 \\
  \multirow{1}{*}{\quad + Ours} &  & 156M & 6 & 6 & 29.13 & +2.13 & 26.93 & +1.05 & +1.59 \\
  \multirow{1}{*}{\quad \textbf{+ More Features} \ding{51}} &  & \textbf{156M} & \textbf{6} & \textbf{6} & \textbf{29.65} & \textbf{+2.65} & 26.89 & +1.01 & \textbf{+1.83} \\
  \hline
  \hline
  \multirow{1}{*}{Deep48} &  & 193M & 48 & 1 & 29.43 & +2.43 & 27.12 & +1.24 & +1.84 \\
  \hdashline[5pt/2.5pt]
  \multirow{1}{*}{Multi-Path2} & & 193M & 24 & 2 & 29.44 & +2.44 & 27.05 & +1.17 & +1.81 \\
  \multirow{1}{*}{\quad \textbf{+ Ours} \ding{51}} &  & \textbf{193M} & \textbf{24} & \textbf{2} & \textbf{29.68} & \textbf{+2.68} & \textbf{27.41} & \textbf{+1.53} & \textbf{+2.11} \\
  \hdashline[5pt/2.5pt] 
  \multirow{1}{*}{Multi-Path4} & & 193M & 12 & 4 & 29.04 & +2.04 & 26.73 & +0.85 & +1.45 \\
  \multirow{1}{*}{\quad + Ours} &  & 193M & 12 & 4 & 29.46 & +2.46 & 27.03 & +1.15 & +1.81 \\
  \multirow{1}{*}{\quad + More Features} &  & 193M & 12 & 4 & 29.56 & +2.56 & 27.00 & +1.12 & +1.84 \\
  \hdashline[5pt/2.5pt] 
  \multirow{1}{*}{Multi-Path8} & & 193M & 6 & 8 & 28.62 & +1.62 & 26.71 & +0.83 & +1.23 \\  
  \multirow{1}{*}{\quad + Ours} &  & 193M & 6 & 8 & 29.21 & +2.21 & 27.07 & +1.19 & +1.70 \\
  \multirow{1}{*}{\quad + More Features} &  & 193M & 6 & 8 & 29.56 & +2.56 & 26.95 & +1.07 & +1.82 \\
  \hline
  \end{tabular}
  \caption{Results on WMT14 En$\rightarrow$De (We mark the best system with \ding{51} under the same number of parameters.
  The original multi-path models with different paths are represented as ``Multi-Path2$\sim$8''.
  Our models with PathNorm and learnable weighted mechanism are represented as ``+ Ours'', our models with more features based on ``+ Ours'' are represented as ``+ More Features''.).}
  \label{tab:ende_base}
\end{table*}

\begin{figure*}[t]
  \centering
  \small
  \begin{tikzpicture}
    \begin{axis}[
      width=4.7cm, height=3.2cm, 
      ybar, ymin=0.6,ymax=1.3,
      grid style=dashed,
      ymajorgrids=true,
      xmajorgrids=true,
      ylabel=$\mathrm{\Delta}_{\text{BLEU}}$,
      xlabel=En$\rightarrow$De,
      ylabel near ticks,
      xlabel near ticks,
      enlarge x limits=0.35,
      bar width=8pt,
      symbolic x coords={Deep,MP,Ours},
      xtick=data,
      nodes near coords style={font=\small}]
      \addplot[draw=lyyred,fill=lyyred!90] coordinates {
      (Deep,1.21)(MP,0.69)(Ours,1.23)};
      \addplot[draw=lyyred,fill=lyyred!90,pattern=north west lines,pattern color=lyyred] coordinates {
      (Deep,1.14)(MP,0.95)(Ours,1.10)};
    \end{axis}
  \end{tikzpicture}
  \begin{tikzpicture}
    \begin{axis}[
      width=4.7cm, height=3.2cm, 
      ybar, ymin=0.5,ymax=0.9,
      grid style=dashed,
      ymajorgrids=true,
      xmajorgrids=true,
      xlabel=En$\rightarrow$Fi,
      ylabel near ticks,
      xlabel near ticks,
      enlarge x limits=0.35,
      bar width=8pt,
      symbolic x coords={Deep,MP,Ours},
      xtick=data,
      nodes near coords style={font=\small}]
      \addplot[draw=lyyblue,fill=lyyblue!90] coordinates {
      (Deep,0.88)(MP,0.64)(Ours,0.80)};
      \addplot[draw=lyyblue,fill=lyyblue!90,pattern=north west lines,pattern color=lyyblue] coordinates {
      (Deep,0.57)(MP,0.63)(Ours,0.62)};
    \end{axis}
  \end{tikzpicture}
  \begin{tikzpicture}
    \begin{axis}[
      width=4.7cm, height=3.2cm, 
      ybar, ymin=0.6,ymax=1.5,
      grid style=dashed,
      ymajorgrids=true,
      xmajorgrids=true,
      xlabel=En$\rightarrow$Lv,
      ylabel near ticks,
      xlabel near ticks,
      enlarge x limits=0.35,
      bar width=8pt,
      symbolic x coords={Deep,MP,Ours},
      xtick=data,
      nodes near coords style={font=\small}]
      \addplot[draw=lyygreen,fill=lyygreen!] coordinates {
      (Deep,0.86)(MP,0.66)(Ours,0.84)};
      \addplot[draw=lyygreen,fill=lyygreen!,pattern=north west lines,pattern color=lyygreen!200] coordinates {
      (Deep,1.44)(MP,1.33)(Ours,1.44)};
    \end{axis}
  \end{tikzpicture}
  \begin{tikzpicture}
    \begin{axis}[
      width=4.7cm, height=3.2cm, 
      ybar, ymin=1.2,ymax=2.0,
      grid style=dashed,
      ymajorgrids=true,
      xmajorgrids=true,
      xlabel=En$\rightarrow$Ru,
      ylabel near ticks,
      xlabel near ticks,
      enlarge x limits=0.35,
      bar width=8pt,
      symbolic x coords={Deep,MP,Ours},
      xtick=data,
      nodes near coords style={font=\small}]
      \addplot[draw=orange,fill=orange!] coordinates {
      (Deep,1.96)(MP,1.50)(Ours,1.90)};
      \addplot[draw=orange,fill=orange!,pattern=north west lines,pattern color=orange!150] coordinates {
      (Deep,1.56)(MP,1.21)(Ours,1.38)};
    \end{axis}
  \end{tikzpicture}
  \begin{tikzpicture}
    \begin{axis}[
      width=4.7cm, height=3.2cm, 
      ybar, ymin=0.7,ymax=1.3,
      grid style=dashed,
      ymajorgrids=true,
      xmajorgrids=true,
      ylabel=$\mathrm{\Delta}_{\text{BLEU}}$,
      xlabel=De$\rightarrow$En,
      ylabel near ticks,
      xlabel near ticks,
      enlarge x limits=0.35,
      bar width=8pt,
      symbolic x coords={Deep,MP,Ours},
      xtick=data,
      nodes near coords style={font=\small}]
      \addplot[draw=lyyred,fill=lyyred!90] coordinates {
        (Deep,1.23)(MP,0.79)(Ours,1.03)};
      \addplot[draw=lyyred,fill=lyyred!90,pattern=north west lines,pattern color=lyyred] coordinates {
        (Deep,0.84)(MP,0.85)(Ours,1.07)};
    \end{axis}
  \end{tikzpicture}
  \begin{tikzpicture}
    \begin{axis}[
      width=4.7cm, height=3.2cm, 
      ybar, ymin=0.1,ymax=0.6,
      grid style=dashed,
      ymajorgrids=true,
      xmajorgrids=true,,
      xlabel=Fi$\rightarrow$En,
      ylabel near ticks,
      xlabel near ticks,
      enlarge x limits=0.35,
      bar width=8pt,
      symbolic x coords={Deep,MP,Ours},
      xtick=data,
      nodes near coords style={font=\small}]
      \addplot[draw=lyyblue,fill=lyyblue!90] coordinates {
      (Deep,0.53)(MP,0.23)(Ours,0.49)};
      \addplot[draw=lyyblue,fill=lyyblue!90,pattern=north west lines,pattern color=lyyblue] coordinates {
      (Deep,0.38)(MP,0.24)(Ours,0.32)};
    \end{axis}
  \end{tikzpicture}
  \begin{tikzpicture}
    \begin{axis}[
      width=4.7cm, height=3.2cm, 
      ybar, ymin=0.0,ymax=1.9,
      grid style=dashed,
      ymajorgrids=true,
      xmajorgrids=true,,
      xlabel=Lv$\rightarrow$En,
      ylabel near ticks,
      xlabel near ticks,
      enlarge x limits=0.35,
      bar width=8pt,
      symbolic x coords={Deep,MP,Ours},
      xtick=data,
      nodes near coords style={font=\small}]
      \addplot[draw=lyygreen,fill=lyygreen!] coordinates {
      (Deep,0.67)(MP,0.09)(Ours,0.58)};
      \addplot[draw=lyygreen,fill=lyygreen!,pattern=north west lines,pattern color=lyygreen!200] coordinates {
      (Deep,1.86)(MP,1.26)(Ours,1.75)};
    \end{axis}
  \end{tikzpicture}
  \begin{tikzpicture}
    \begin{axis}[
      width=4.7cm, height=3.2cm, 
      ybar, ymin=1.4,ymax=1.8,
      grid style=dashed,
      ymajorgrids=true,
      xmajorgrids=true,,
      xlabel=Ru$\rightarrow$En,
      ylabel near ticks,
      xlabel near ticks,
      enlarge x limits=0.35,
      bar width=8pt,
      symbolic x coords={Deep,MP,Ours},
      xtick=data,
      nodes near coords style={font=\small}]
      \addplot[draw=orange,fill=orange!] coordinates {
      (Deep,1.73)(MP,1.69)(Ours,1.80)};
      \addplot[draw=orange,fill=orange!,pattern=north west lines,pattern color=orange!150] coordinates {
      (Deep,1.62)(MP,1.41)(Ours,1.59)};
    \end{axis}
  \end{tikzpicture}
  \caption{Comparisons of different systems on WMT17 tasks
  (The figure shows $\mathrm{\Delta}_{\text{BLEU}}$ of each system exceeds the baseline model. 
  Deep, MP, and Ours denote results of deep models, original multi-path models, and our multi-path models.
  Blocks with/without dotted lines represent results on validation/test set.
  Different colors {\color{lyyred!90}{$\blacksquare$}}{\color{lyyblue!90}{$\blacksquare$}}{\color{lyygreen}{$\blacksquare$}}{\color{orange}{$\blacksquare$}} denote tasks of En$\leftrightarrow$De, En$\leftrightarrow$Fi, En$\leftrightarrow$Lv and En$\leftrightarrow$Ru.).}
  \label{fig:comparison_system}
\end{figure*}
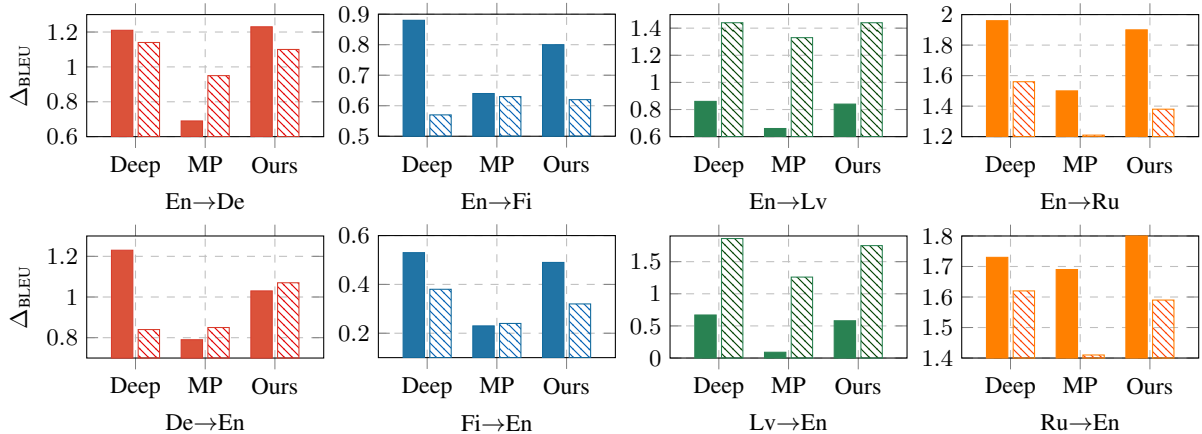

\subsection{Setup}

We evaluate our methods on 12 machine translation tasks (6 datasets $\times$ 2 translation directions each), including WMT14 English$\leftrightarrow$\{German, French\} (En$\leftrightarrow$\{De, Fr\}) and WMT17 English$\leftrightarrow$\{German, Finnish, Latvian, Russian\} (En$\leftrightarrow$\{De, Fi, Lv, Ru\}).
The statistics of all datasets are shown in \tab{tab:data}.

\textbf{Datasets.}
For the En$\leftrightarrow$De tasks (4.5M pairs), we choose \emph{newstest-2013} as the validation set and \emph{newstest-2014} as the test set. 
We share the source and target vocabularies. 
For the En$\leftrightarrow$Fr tasks (35M pairs), we validate the system on the combination of newstest-2012 and newstest-2013, and test it on newstest-2014.
We use the concatenation of all available preprocessed validation sets in WMT17 datasets as our validation set.
All WMT datasets are provided within the official website\footnote{http://statmt.org/}.

For all datasets, we tokenize every sentence using the script in the Moses\footnote{https://github.com/moses-smt/mosesdecoder/blob/maste\\r/scripts/tokenizer/tokenizer.perl} toolkit and segment every word into subword units using Byte-Pair Encoding \cite{DBLP:conf/acl/SennrichHB16a}. 
The number of the BPE merge operations is set to 32K in all these tasks. 
In addition, we remove sentences with more than 250 subword units \cite{DBLP:conf/acl/XiaoZZL12} and evaluate the results using multi-bleu.perl\footnote{https://github.com/moses-smt/mosesdecoder/blob/maste\\r/scripts/generic/multi-bleu.perl}.

\textbf{Models.}
Our baseline system is based on the open-source implementation of the Transformer model presented in \citet{DBLP:conf/naacl/OttEBFGNGA19}'s work.
For all machine translation tasks, we construct baseline models with the Transformer-base and Transformer-deep \cite{DBLP:conf/acl/WangLXZLWC19} settings. 
All baseline systems consist of a 6-layer encoder and a 6-layer decoder, except that the Transformer-deep encoder has 12$\sim$48 layers (depth) \cite{DBLP:journals/corr/abs-2012-13866}. 
The embedding size is set to 512 for both the Transformer-base and deep. 
The FFN hidden size equals 4$\times$ embedding size in all settings. 
As for the multi-path Transformer models, except for the number of paths, all other model hyperparameters are the same as the baseline models.
The multi-path models in this paper consist of 2$\sim$8 paths.

\textbf{Training Details.}
For training, we use Adam optimizer with $\beta_1=0.9$ and $\beta_2=0.997$.
For the Transformer-base setting, we adopt the inverse square root learning rate schedule with 8,000 warmup steps and $0.001$ learning rate.
For the Transformer-deep and multi-path settings, we adopt the inverse square root learning rate schedule with 16,000 warmup steps and $0.002$ learning rate.
The training batch size of 4,096 is adopted in the base setting, and 8,192 is adopted in the deep and multi-path settings. 
All experiments are done on 8 NVIDIA TITIAN V GPUs with mixed-precision training \cite{DBLP:conf/iclr/MicikeviciusNAD18}. 
All results are reported based on the model ensembling by averaging the last 5 checkpoints.

\subsection{Results}

\begin{table}[t!]
  \centering
  \small
  \renewcommand\arraystretch{1.15}
  \setlength{\tabcolsep}{0.07cm}
  \begin{tabular}{c|l|r|c|c|c|c}
  \hline
  &
  \multicolumn{1}{l|}{System} &
  Params &
  \multicolumn{1}{c|}{D-P} &
  \multicolumn{1}{c|}{Test} &
  \multicolumn{1}{c|}{Valid} &
  \multicolumn{1}{c}{$\mathrm{\Delta}_{\text{Average}}$} \\
  \hline
  \multirow{7}*{\rotatebox{90}{WMT14 De-En}} & Baseline & 62M & 6-1 & 30.50 & 30.34 & +0.00 \\
  \cline{2-7}
  & \textbf{Deep24} \ding{51} & \textbf{118M} & \textbf{24-1} & 31.92 & \textbf{31.37} & \textbf{+1.23} \\
  \cdashline{2-7}[5pt/2.5pt]
  & \multirow{1}{*}{Multi-Path2} & 118M & 12-2 & 31.62 & 31.00 & +0.89 \\
  & \multirow{1}{*}{\quad \textbf{+ Ours} \ding{51}} & \textbf{118M} & \textbf{12-2} & \textbf{32.00} & 31.29 & \textbf{+1.23} \\
  \cdashline{2-7}[5pt/2.5pt]
  & \multirow{1}{*}{Multi-Path4} & 118M & 6-4 & 31.52 & 30.80 & +0.74 \\
  & \multirow{1}{*}{\quad + Ours} & 118M & 6-4 & 31.85 & 30.95 & +0.98 \\
  & \multirow{1}{*}{\quad + More Features} & 118M & 6-4 & 31.89 & 31.03 & +1.04 \\
  \hline
  \hline
  \multirow{7}*{\rotatebox{90}{WMT14 En-Fr}} & Baseline & 111M & 6-1 & 40.82 & 46.80 & +0.00 \\
  \cline{2-7}
  & Deep24 & 168M & 24-1  & 42.40 & 48.41 & +1.60 \\
  \cdashline{2-7}[5pt/2.5pt]
  & \multirow{1}{*}{Multi-Path2} & 168M & 12-2 & 42.40 & 48.37 & +1.58 \\
  & \multirow{1}{*}{\quad \textbf{+ Ours} \ding{51}} & \textbf{168M} & \textbf{12-2} & \textbf{42.44} & \textbf{48.45} & \textbf{+1.64} \\
  \cdashline{2-7}[5pt/2.5pt]
  & \multirow{1}{*}{Multi-Path4} & 168M & 6-4 & 41.76 & 47.93 & +1.04 \\
  & \multirow{1}{*}{\quad + Ours} & 168M & 6-4 & 41.90 & 48.19 & +1.24 \\
  & \multirow{1}{*}{\quad + More Features} & 168M & 6-4 & 42.32 & 48.37 & +1.54 \\
  \hline
  \hline
  \multirow{7}*{\rotatebox{90}{WMT14 Fr-En}} & Baseline & 111M & 6-1 & 36.33 & 47.03 & +0.00 \\
  \cline{2-7}
  & Deep24 & 168M & \multicolumn{4}{c}{\textbf{loss exploding \ding{55}}} \\
  \cdashline{2-7}[5pt/2.5pt]
  & \multirow{1}{*}{Multi-Path2} & 168M & 12-2 & 38.19 & 48.02 & +1.43 \\
  & \multirow{1}{*}{\quad \textbf{+ Ours} \ding{51}} & \textbf{168M} & \textbf{12-2} & 38.24 & \textbf{48.45} & \textbf{+1.67} \\
  \cdashline{2-7}[5pt/2.5pt]
  & \multirow{1}{*}{Multi-Path4} & 168M & 6-4 & 37.94 & 48.18 & +1.38 \\
  & \multirow{1}{*}{\quad + Ours} & 168M & 6-4 & 38.20 & 48.30 & +1.57 \\
  & \multirow{1}{*}{\quad + More Features} & 168M & 6-4 & \textbf{38.26} & 48.36 & +1.63 \\
  \hline
  \end{tabular}
  \caption{Results on other WMT14 tasks (We mark the best system with \ding{51}, \ding{55} means that the model cannot continue training due to the gradient exploding problem.).}
  \label{tab:wmt14_reverse}
\end{table}

\tab{tab:ende_base} and \tab{tab:wmt14_reverse} show the results of different systems on WMT14 En$\leftrightarrow$De and WMT14 En$\leftrightarrow$Fr.
In all tasks, the original multi-path models can not perform as well as the deep models, which proves that model depth does play a crucial role in performance.
However, our multi-path models can achieve similar or even better results than the deep models.
On the En$\rightarrow$De dataset, our best multi-path system achieves 0.12/0.32/0.28/0.25 higher BLEU points than the deep model when the number of parameters is set to 80/118/156/193 megabytes.
It shows the potential of the multi-path models and proves that model width is as important as model depth.
Under the same model depth, multi-path models with more features significantly perform better than the original multi-path models, which demonstrates the effectiveness of our proposed method in Section \ref{sec:methods}.
Note that in our method of generating more features, there will be $C_n^{n-1}=n$ new features.
In a 2-path model, since there are only 2 paths, no new features will be generated.

Experiments on En$\rightarrow$Fr, De$\rightarrow$En and Fr$\rightarrow$En also show the competitive performance of the multi-path Transformer models.
Note that on Fr$\rightarrow$En task, in the mix-precision training process of the 24-layer model, we met the problem of loss exploding.
It means that the minimum loss scale (0.0001 in the fairseq fp16 optimizer\footnote{https://github.com/pytorch/fairseq/blob/v0.6.2/fairseq/opt\\im/fp16\_optimizer.py}) has been reached and the loss is probably exploding.
We further validate our conclusions on 8 WMT17 tasks, including En$\leftrightarrow$\{De, Fi, Lv, Ru\}. 
Experiments in \fig{fig:comparison_system} show a similar phenomenon and
further verify that,
instead of indefinitely stacking more layers, we should pay more attention to wider structures, such as the multi-path models.

\section{Analysis}

\subsection{Shallower Networks}

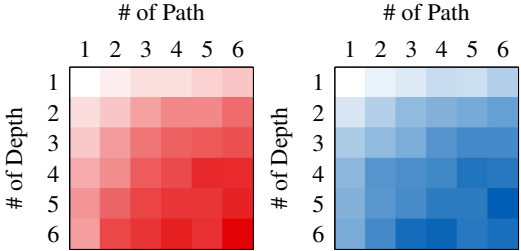
\begin{figure}[!t]
  \centering
  \begin{tikzpicture}
    \begin{axis}[
        width=0.52\linewidth,height=0.52\linewidth,
        view={0}{90},
        enlargelimits=false,
        ymin=0.5,ymax=6.5,
        xmin=0.5,xmax=6.5,
        xlabel={\# of Path},
        ylabel={\# of Depth},
        ytick={1,2,3,4,5,6},
        yticklabels={1,2,3,4,5,6},
        xtick={1,2,3,4,5,6},
        xticklabels={1,2,3,4,5,6},
        label style={font=\small},
        every tick label/.append style={font=\small},
        xticklabel pos=upper,
        xtick style={draw=none},
        ytick style={draw=none},
        colormap={bluepurple}{color=(white) color=(lyyred)},
    ]
      \addplot3[matrix plot] table [meta=value] {
        x y value
        1 1 25.31
        1 2 25.88
        1 3 26.24
        1 4 26.74
        1 5 27.12
        1 6 26.95

        2 1 25.59
        2 2 26.27
        2 3 26.99
        2 4 27.24
        2 5 27.92
        2 6 28.44

        3 1 25.86
        3 2 26.90
        3 3 27.66
        3 4 28.10
        3 5 28.49
        3 6 28.75

        4 1 25.86
        4 2 27.35
        4 3 28.00
        4 4 28.38
        4 5 28.74
        4 6 29.11

        5 1 26.11
        5 2 27.33
        5 3 28.15
        5 4 28.95
        5 5 28.74
        5 6 28.84

        6 1 26.30
        6 2 27.82
        6 3 28.28
        6 4 28.92
        6 5 29.07
        6 6 29.65
      };
    \end{axis}
  \end{tikzpicture}
  \begin{tikzpicture}
    \begin{axis}[
        width=0.52\linewidth,height=0.52\linewidth,
        view={0}{90},
        enlargelimits=false,
        ymin=0.5,ymax=6.5,
        xmin=0.5,xmax=6.5,
        xlabel={\# of Path},
        ylabel={\# of Depth},
        ytick={1,2,3,4,5,6},
        yticklabels={1,2,3,4,5,6},
        xtick={1,2,3,4,5,6},
        xticklabels={1,2,3,4,5,6},
        label style={font=\small},
        every tick label/.append style={font=\small},
        xticklabel pos=upper,
        xtick style={draw=none},
        ytick style={draw=none},
        colormap={bluepurple}{color=(white) color=(lyyblue)},
    ]
      \addplot3[matrix plot] table [meta=value] {
        x y value
        1 1 24.48
        1 2 24.89
        1 3 25.37
        1 4 25.67
        1 5 25.78
        1 6 25.88

        2 1 24.71
        2 2 25.27
        2 3 25.63
        2 4 26.25
        2 5 26.22
        2 6 26.43

        3 1 24.84
        3 2 25.64
        3 3 25.85
        3 4 26.36
        3 5 26.41
        3 6 26.94

        4 1 25.06
        4 2 25.78
        4 3 26.26
        4 4 26.47
        4 5 26.68
        4 6 27.05

        5 1 25.02
        5 2 25.90
        5 3 26.44
        5 4 26.82
        5 5 26.69
        5 6 26.73

        6 1 25.29
        6 2 26.08
        6 3 26.42
        6 4 26.73
        6 5 27.17
        6 6 26.89
      };
    \end{axis}
  \end{tikzpicture}
  \caption{Results of shallower networks with different depths and paths on WMT14 En$\rightarrow$De (Darker color means better performance.).}
  \label{fig:shallower}
\end{figure}

In this section, we study the performance of our multi-path structure in shallower networks.
\fig{fig:shallower} shows the results of Transformer models with different numbers of depths and paths.
When the model is relatively shallower, increasing the number of paths will produce slightly worse performance than increasing the number of depths (e.g., the 1-layer 6-path model vs. the 6-layer 1-path model). 
When the model is deeper, increasing the number of paths has a greater advantage (e.g., the 2-layer 5-depth model vs. the 5-layer 2-depth models).
In most instances, changing the depth and path have almost the same effect on model performance.

\begin{table}[!t]
  \centering
  \small
  \renewcommand\arraystretch{1.11}
  \setlength{\tabcolsep}{2.9mm}{
    \begin{tabular}{l|r|c|c}
      \hline
      \multicolumn{1}{c|}{\multirow{2}{*}{System}}
      &
      \multicolumn{1}{c|}{\multirow{2}{*}{Params}} &
      \multicolumn{2}{c}{BLEU} \\
      \cline{3-4}
      & &   
      \multicolumn{1}{c|}{Test} & 
      \multicolumn{1}{c}{Valid} \\
      \hline
      \multirow{1}{*}{Baseline} & 62M & 27.00 & 25.88 \\
      \hline
      \multirow{1}{*}{+ Multi-Path} & 156M & 28.87 & 26.58 \\
      \multirow{1}{*}{+ PathNorm} & 156M & 28.72 & 26.28 \\
      \multirow{1}{*}{+ Learnable Weights} & 156M & 29.13 & 26.93 \\
      \multirow{1}{*}{\quad - PathNorm} & 156M & 28.86 & 26.58 \\
      \multirow{1}{*}{\textbf{+ More Features}} & 156M & \textbf{29.65} & \textbf{26.89} \\
      \hline
    \end{tabular}
  \caption{Ablation study on WMT14 En$\rightarrow$De.}
  \label{tab:ablation}
  }
\end{table}

\subsection{Ablation Study}

\tab{tab:ablation} summarizes and compares the contributions of each part described in Section \ref{sec:methods}. 
Each row of \tab{tab:ablation} is the result of applying the current part to the system obtained in the previous row. 
This way helps to illustrate the compound effect of these parts.
Here we adopt the 6-layer 6-path model for study.
In the first two rows, different paths in the same layer are added with the fixed weights (1/6 in the + Multi-Path model and 1/$\sqrt{6}$ in the + PathNorm model).
We can see that the + Multi-Path model significantly surpasses the baseline model.
However, the + PathNorm model performs slightly worse than the + Multi-Path model.
In order to verify the importance of PathNorm in our method, we conduct an additional experiment to use learnable weights alone (- PathNorm).
As can be seen in \tab{tab:ablation}, neither the learnable weights (- PathNorm) nor PathNorm (+ PathNorm) works well alone, which verifies the importance of the combination of learnable weights and PathNorm (+ Learnable Weights) in our method.

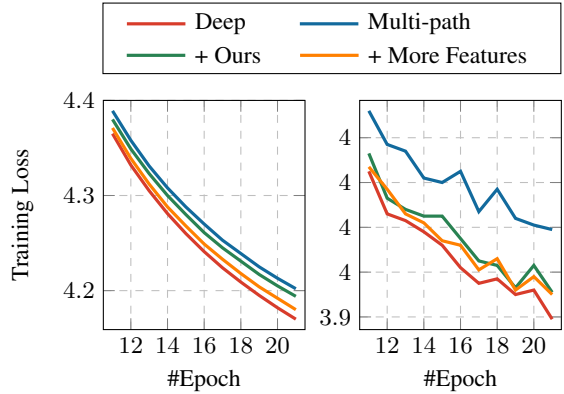
\begin{figure}[t!]
  \hspace{1.02cm}
  \tikz {
      \small
      \legendmargin=0.04\linewidth
      \legendwidth=0.1\linewidth,
      \legendsep=0.05\linewidth
      \coordinate (start) at (0,0);
      \draw[lyyred,very thick] ([xshift=\legendmargin]start.east) -- +(\legendwidth,0) node[black,right] (l1) {Deep};
      \draw[lyyblue,very thick] ([xshift=0.076\linewidth]l1.east) -- +(\legendwidth,0) node[black,right] (l2) {Multi-path};
      \draw[lyygreen,very thick] ([xshift=-0.16\linewidth,yshift=-0.02\linewidth]l1.south) -- +(\legendwidth,0) node[black,right] (l3) {+ Ours};
      \draw[orange,very thick] ([xshift=0.05\linewidth]l3.east) -- +(\legendwidth,0) node[black,right] (l4) {+ More Features};
      \coordinate (end) at ([xshift=\legendmargin+0pt]l4.east);
      \begin{pgfonlayer}{background}
      \node[rectangle,draw,inner sep=0.2pt] [fit = (start) (l1) (l2) (l3) (l4) (end)] {};
      \end{pgfonlayer}
  }
  \\[3pt]
  \centering
  \small
    \begin{tikzpicture}
      \begin{axis}[
          width=0.55\linewidth,height=0.6\linewidth,
          yticklabel style={/pgf/number format/fixed,/pgf/number format/precision=1},
          ylabel={Training Loss},
          ylabel near ticks,
          xlabel={\#Epoch},
          xlabel near ticks,
          enlargelimits=0.05,
          xtick distance=2,
          xmajorgrids=true,
          ymajorgrids=true,
          grid style=dashed,
          every tick label/.append style={font=\small},
          label style={font=\small},
          ylabel style={yshift=5pt},
          legend style={font=\small,inner sep=3pt},
          legend image post style={scale=1},
          legend columns=2,
          legend cell align={left},
      ]
      \addplot [lyyred,very thick] coordinates {
          (11,4.365) (12,4.332) (13,4.305) (14,4.281) (15,4.260) (16,4.241) (17,4.224) (18,4.209) (19,4.195) (20,4.182) (21,4.170)
      };
      \addplot [lyyblue,very thick] coordinates {
          (11,4.389) (12,4.358) (13,4.331) (14,4.308) (15,4.288) (16,4.270) (17,4.253) (18,4.239) (19,4.225) (20,4.213) (21,4.202)
      };
      \addplot [lyygreen,very thick] coordinates {
          (11,4.380) (12,4.349) (13,4.323) (14,4.300) (15,4.280) (16,4.261) (17,4.245) (18,4.231) (19,4.217) (20,4.205) (21,4.194)
      };
      \addplot [orange,very thick] coordinates {
          (11,4.371) (12,4.339) (13,4.312) (14,4.288) (15,4.268) (16,4.249) (17,4.233) (18,4.218) (19,4.204) (20,4.192) (21,4.180)
      };
      \end{axis}
    \end{tikzpicture}
    \begin{tikzpicture}
      \begin{axis}[
          width=0.55\linewidth,height=0.6\linewidth,
          yticklabel style={/pgf/number format/fixed,/pgf/number format/precision=1},
          ylabel near ticks,
          xlabel={\#Epoch},
          xlabel near ticks,
          enlargelimits=0.05,
          xtick distance=2,
          xmajorgrids=true,
          ymajorgrids=true,
          grid style=dashed,
          every tick label/.append style={font=\small},
          label style={font=\small},
          ylabel style={yshift=5pt},
          legend style={font=\small,inner sep=3pt},
          legend image post style={scale=1},
          legend columns=2,
          legend cell align={left},
      ]
      \addplot [lyyred,very thick] coordinates {
          (11,4.005) (12,3.986) (13,3.983) (14,3.978) (15,3.972) (16,3.962) (17,3.955) (18,3.957) (19,3.950) (20,3.952) (21,3.939)
      };
      \addplot [lyyblue,very thick] coordinates {
          (11,4.032) (12,4.017) (13,4.014) (14,4.002) (15,4.000) (16,4.005) (17,3.987) (18,3.997) (19,3.984) (20,3.981) (21,3.979)
      };
      \addplot [lyygreen,very thick] coordinates {
          (11,4.013) (12,3.993) (13,3.988) (14,3.985) (15,3.985) (16,3.975) (17,3.965) (18,3.963) (19,3.953) (20,3.963) (21,3.951)
      };
      \addplot [orange,very thick] coordinates {
          (11,4.007) (12,3.997) (13,3.986) (14,3.982) (15,3.974) (16,3.972) (17,3.961) (18,3.966) (19,3.952) (20,3.958) (21,3.950)
      };
      \end{axis}
    \end{tikzpicture}
  \hspace{\fill}
  \caption{Loss vs. the number of epochs on WMT14 En-De (The left figure plots the training losses, the right figure plots the validation losses.).}
  \label{fig:curve}
\end{figure}

\subsection{Training Study}

We plot the training and validation loss curves of different systems with the same number of parameters, including the deep model, the original multi-path model (Multi-Path) and our model without/with more features (+ Ours/+ More Features), for studying their convergence.
All these four systems have been shown in \tab{tab:ende_base}.
We can see that all systems converge stably.
The original multi-path model has a higher loss than other models in both the training and validation sets and it does perform the worst.
The deep model has the lowest loss, but the performance is close to + Ours and + More Features, which means that the loss cannot absolutely reflect the model performance.

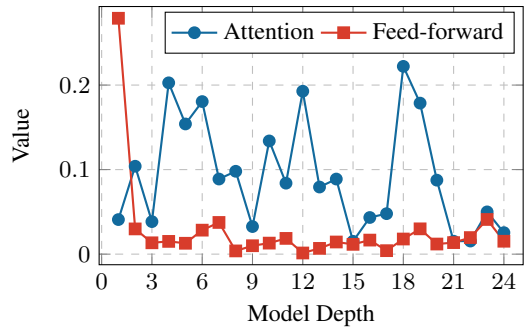
\begin{figure}
  \centering
  \small
  \begin{tikzpicture}
      \begin{axis}[
          width=0.93\linewidth,height=0.65\linewidth,
          yticklabel style={/pgf/number format/fixed,/pgf/number format/precision=1},
          ylabel={Value},
          ylabel near ticks,
          xlabel={Model Depth},
          xlabel near ticks,
          enlargelimits=0.05,
          xtick distance=3,
          xmajorgrids=true,
          ymajorgrids=true,
          grid style=dashed,
          every tick label/.append style={font=\small},
          label style={font=\small},
          ylabel style={yshift=5pt},
          legend style={font=\small,inner sep=2pt},
          legend image post style={scale=1},
          legend cell align={left},
          legend columns=2,
      ]
          \addplot [lyyblue,thick,mark=*] coordinates {
              (1,0.0171/0.4183) (2,0.2103/2.0224) (3,0.0314/0.8162) 
              (4,0.2133/1.0525) (5,0.1794/1.1650) (6,0.3166/1.7542) (7,0.1556/1.7514) (8,0.1438/1.4670) (9,0.0429/1.3149) (10,0.1574/1.1754) (11,0.1191/1.4189) (12,0.2875/1.4917) (13,0.1250/1.5748) (14,0.1165/1.3115) (15,0.0233/1.5359) (16,0.0530/1.2228) (17,0.0726/1.5164) (18,0.3643/1.6401) (19,0.2447/1.3697) (20,0.0857/0.9793) (21,0.0216/1.4108) (22,0.0222/1.4334) (23,0.0603/1.2061) (24,0.0309/1.2154)
          };
          \addlegendentry{Attention}
          \addplot [lyyred,thick,mark=square*] coordinates {
              (1,0.4422/1.5848) (2,0.0500/1.6758) (3,0.0226/1.6846) 
              (4,0.0241/1.5981) (5,0.0191/1.4849) (6,0.0429/1.5119) (7,0.0607/1.6195) (8,0.0058/1.5204) (9,0.0160/1.6090) (10,0.0199/1.5257) (11,0.0283/1.5205) (12,0.0021/1.5429) (13,0.0103/1.5141) (14,0.0225/1.5665) (15,0.0181/1.5639) (16,0.0267/1.6067) (17,0.0064/1.5564) (18,0.0301/1.6823) (19,0.0516/1.7243) (20,0.0212/1.7730) (21,0.0257/1.8829) (22,0.0367/1.8727) (23,0.0756/1.8436) (24,0.0262/1.7158)
          };
          \addlegendentry{Feed-forward}
      \end{axis}
  \end{tikzpicture}
  \caption{The diversities among different paths vs. the model depth on WMT14 En$\rightarrow$De.}
  \label{fig:weights}
\end{figure}

\subsection{Learnable Weights}

As the learnable $\alpha$ is considered to be a way of measuring the importance of different paths,
the difference of $\alpha$ from different paths can be seen as the diversities among these paths.
\fig{fig:weights} studies the behavior of $\alpha$, the solid lines denote the absolute value of the difference of $\alpha$.
Here we adopt the 24-layer 2-path Transformer system to study $\alpha$ in deep models, we let $\left|d \right|=\frac{\left| \alpha_1-\alpha_2 \right|}{\left| \alpha_1+\alpha_2 \right|}$ denote the above mentioned absolute value of difference.

As can be seen in \fig{fig:weights}, either in the attention layer or the feed-forward layer, $\left|d \right|$ changes significantly in different model depths.
In the feed-forward layer, except for the first and last several layers (e.g., 1, 22, and 23), the value of $\left|d \right|$ is smaller than the attention layer, which reflects from the side that the diversity of the feed-forward layer is smaller than the attention layer.
In the attention layer, the value of $\left|d \right|$ is larger in the middle layers (e.g., from 4 to 20), indicating that more model diversities can be learned in the middle layers than the bottom and top layers.

\subsection{Training Efficiency} 
\label{subsec:training_efficiency}

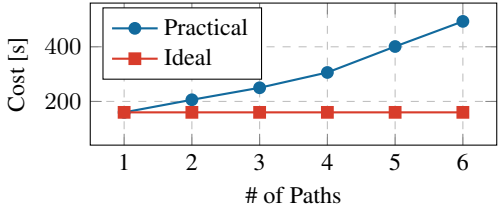
\begin{figure}[t!]
  \centering
  \small
  \begin{tikzpicture}
      \begin{axis}[
          width=0.9\linewidth,height=0.45\linewidth,
          yticklabel style={/pgf/number format/fixed,/pgf/number format/precision=1},
          ymin=100,ymax=500,
          enlarge y limits=0.15,
          ylabel={Cost [s]},
          ylabel near ticks,
          xlabel={\# of Paths},
          xlabel near ticks,
          symbolic x coords={1,2,3,4,5,6},
          xmajorgrids=true,
          ymajorgrids=true,
          grid style=dashed,
          xtick=data,
          every tick label/.append style={font=\small},
          label style={font=\small},
          legend pos=north west,
          legend columns=1,
          legend cell align={left},
        ]
          \addplot [lyyblue,thick, mark=*] coordinates {
              (1,160) (2,206) (3,250) (4,306) (5,401) (6,493)
          };
          \addlegendentry{Practical}
          \addplot [lyyred,thick, mark=square*] coordinates {
              (1,160) (2,160) (3,160) (4,160) (5,160) (6,160)
          };
          \addlegendentry{Ideal}
      \end{axis}
  \end{tikzpicture}
  \caption{Training cost vs. the number of paths on WMT14 En$\rightarrow$De.}
  \label{fig:training_efficiency}
\end{figure}

Here we record the computation times required per 100 training steps of different models.
To exclude the influence of data transfer, we train these models on a single GPU.
Since each path is computed independently, the multi-path structure adopted in this paper has the inherent advantage of high computational parallelism.
However, due to the limitations of related computational libraries, this kind of model does not achieve its ideal efficiency as can be seen in \fig{fig:training_efficiency}.
As one type of model structure with great potential, the multi-path model should get more attention from us, and the related computational libraries should also be completed.


\section{Discussion}

Model depth or width which is more important becomes a hot topic in recent years \cite{DBLP:conf/iclr/NguyenRK21,DBLP:journals/corr/abs-2202-03841,DBLP:conf/colt/EldanS16,DBLP:conf/nips/LuPWH017,DBLP:conf/recsys/Cheng0HSCAACCIA16}.
In general, one model can benefit more from increasing the depth \cite{DBLP:conf/nips/KrizhevskySH12,DBLP:journals/corr/SimonyanZ14a,DBLP:conf/cvpr/SzegedyLJSRAEVR15}, the reasons can be summarized as follows: 
1) Expressivity, deep models have better non-linear expressivity to learn more complex transformations.
2) Efficiency, both the number of parameters and the computational complexity will be changed quadratically corresponding to the model width (referring to scaling the matrix dimensions) while linearly with the model depth, thus the cost of increasing width is often much higher than that of depth.

For tasks such as computer vision, the model depth can even reach hundreds or thousands of layers \cite{DBLP:conf/cvpr/HeZRS16,DBLP:journals/corr/ZagoruykoK17}.
For the Transformer model,
\citet{DBLP:journals/corr/abs-2203-00555} even train a Transformer model with 1,000 layers.
However, the training process of deep models is not as simple as scaling the number of layers.
When the model becomes too deep, the degradation problem caused by the back propagation will be exposed \cite{DBLP:conf/cvpr/HeZRS16}.

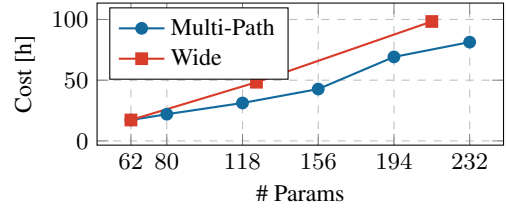
\begin{figure}[t!]
  \centering
  \small
  \begin{tikzpicture}
      \begin{axis}[
          width=0.9\linewidth,height=0.45\linewidth,
          yticklabel style={/pgf/number format/fixed,/pgf/number format/precision=1},
          ymin=10,ymax=100,
          enlarge y limits=0.15,
          ylabel={Cost [h]},
          ylabel near ticks,
          xlabel={\# Params},
          xlabel near ticks,
          xmajorgrids=true,
          ymajorgrids=true,
          grid style=dashed,
          xtick=data,
          every tick label/.append style={font=\small},
          label style={font=\small},
          legend pos=north west,
          legend columns=1,
          legend cell align={left},
        ]
          \addplot [lyyblue,thick, mark=*] coordinates {
              (62,17.22) (80,21.98) (118,31.22) (156,42.70) (194,69.16) (232,81.34)
          };
          \addlegendentry{Multi-Path}
          \addplot [lyyred,thick, mark=square*] coordinates {
            (62,17.22) (125,48.33) (213,98.33)
          };
          \addlegendentry{Wide}
      \end{axis}
  \end{tikzpicture}
  \caption{Training costs of different systems on WMT14 En$\rightarrow$De.}
  \label{fig:training_cost}
\end{figure}

To seek new solutions to further improve large-scale neural networks, here we adopt the parameter-efficient multi-path structure.
As shown in \fig{fig:efficiency}, the multi-path models significantly outperform wide models that scale the matrix dimensions.
From \fig{fig:training_efficiency} and \fig{fig:training_cost} we can see that, although the multi-path model does not achieve its ideal efficiency because of computational libraries support, it still takes less training cost than wide models.
The above discussions show that multi-path is a better option to broaden the model width, and the width of one model is as important as its depth for the purpose of improving capacity.

\section{Conclusion}

In this work, we construct a sublayer-level multi-path structure to study how model width affects the Transformer model.
To better fuse features extracted from different paths, three additional operations mentioned in Section \ref{sec:methods} are introduced.
The experimental results on 12 machine translation benchmarks validate our point of view that, instead of indefinitely stacking more layers, there should be a balance between the model depth and width to train a better large-scale Transformer.

\section*{Acknowledgments}

This work was supported in part by the National Science Foundation of China (Nos. 61876035 and 61732005), the China HTRD Center Project (No. 2020AAA0107904), Yunnan Provincial Major Science and Technology Special Plan Projects (Nos. 202002AD080001 and 202103AA080015), National Frontiers Science Center for Industrial Intelligence and Systems Optimization (Northeastern University, China. No. B16009) and the Fundamental Research Funds for the Central Universities.
The authors would like to thank anonymous reviewers for their comments.

\section*{Limitations}

For the limitation of our work, we will discuss it from three aspects.

\textbf{Non-Ideal Training Efficiency.}
As discussed in Section \ref{subsec:training_efficiency}, although the multi-path structure adopted in this paper has an inherent advantage of high computational parallelism, the training efficiency of this kind of model does not achieve its theoretical height.
As one type of model structure with great potential, the multi-path network should get more attention from us, and the related computational libraries should also be completed.

\textbf{Non-Optimal Hyperparameters.}
Just like the training hyperparameters are quite different among Transformer-base, big and deep systems, the optimal hyperparameters for models with different depths and widths tend to be different.
However, due to the limited computing resources, we do not tune but choose the same hyperparameters as the deep models, which may lead to a non-optimal setting.

\textbf{Very Large-Scale Networks.}
Limited by the hardware and memory resources, we did not explore very large models with much more layers and paths. 
All we can do here is provide insights about how to choose a better combination of model depth and width with limited resources.


\bibliography{anthology,custom}
\bibliographystyle{acl_natbib}

\end{document}